# A Novel Dataset Towards Extracting Virus-Host Interactions

**Rasha Alshawi, Atriya Sen**
University of New Orleans
**Nathan S. Upham, Beckett Sterner**
Arizona State University

## Abstract

We describe a novel dataset for the automated recognition of named taxonomic and other entities relevant to the association of viruses with their hosts. We further describe some initial results using pre-trained models on the named-entity recognition (NER) task on this novel dataset. We propose that our dataset of manually annotated abstracts now offers a Gold Standard Corpus for training future NER models in the automated extraction of host-pathogen detection methods from scientific publications, and further explain how our work makes first steps towards predicting the important human health-related concept of viral spillover risk automatically from the scientific literature.

## 1 Introduction

The pace of novel zoonotic diseases is increasing globally (Han et al. 2016), but much of our knowledge about the geography and hosts of zoonotic diseases remains locked in the texts of published scientific articles (Upham et al. 2021). Published studies typically apply one or more methods for pathogen detection in animal hosts, including antibody tests, polymerase chain reaction (PCR) tests, whole genome sequencing, or live pathogen isolation. Similarly, the host species might be identified morphologically or using PCR. These methods of detecting host-pathogen interactions vary in precision and in what they tell us about the ecological relationship being observed; most critically, whether the animal host is a reservoir for pathogen replication and transmission, or else a more transient host.

Distinguishing the confidence in host-pathogen data according to type of detection method has been shown to significantly improve models predicting zoonotic disease risk in rodents (Mull et al. 2021). However, the information required to incorporate detection method as a variable in zoonotic disease risk models is rarely available from current host-pathogen databases or article metadata, with the important exception of (Olival et al., 2017), which we explore further here. Therefore, we highlight that Named Entity Recognition (NER) methods have great potential to extract in a high-throughput manner the detection methods for host-pathogen interactions, enabling advances in scientific understanding of how and why zoonotic diseases emerge.

More generally, the Information Extraction (IE) challenges in the NER task on biological scientific articles are highly similar to those in other domains, such as astrophysics, as exemplified by the *DEAL: Detecting Entities in the Astrophysics Literature* (DEAL, 2022) competition. It is true of many domains that there is a diversity of naming practices, rampant ambiguity, and a highly dynamic vocabulary. We therefore envision our approach piloted here on host-pathogen literature to be highly generalizable to other scientific domains.

A description of the Sections of this paper follows. Section 2 provides a description of the novel dataset we introduce, which comprises of manually annotated abstracts of scientific publications. Section 3 describes a pretrained deep neural model for the NER task that uses a transformer-based architecture. Section 4 provides quantitative results, and finally, Section 5 discusses various avenues for future work and potential impact.

## 2 Dataset

Our virus dataset was built by manually collecting 1104 articles reporting virus detection results for mammal hosts. The articles were selected from the dataset collected and analyzed as part of a systematic literature review of all known viruses with mammal hosts, as reported in (Olival et al., 2017).

Their review searched the Web of Science, Google Scholar, and PubMed for articles published between 1940 and 2015 that mentioned each of 586 virus species identified as having mammal hosts by the International Committee on Taxonomy of Viruses, 8th Edition (Fauquet et al. 2005). They excluded articles reporting results from experimental infections, zoos, or captive breeding facilities as well as domesticated and peri-domestic mammal species (specifically *Mus musculus* and *Rattus norvegicus*). The final list of articles they analyzed is available online as part of the paper's supplementary data in the "references.txt" file (https://zenodo.org/record/807517). Since the supplementary data did not include article abstracts, we searched by article title in the PubMed search engine (PubMed,2022), and with our additions, the dataset we report here represents a substantial fraction of all published articles reporting viruses detected in wild mammals.

524 of these abstracts were preprocessed and manually annotated in the form of Gold Standard Corpus (GSC) for the Name Entity Recognition (NER) task. A GSC is a collection of manually annotated documents. NER annotation requires the inclusion of at least the left and right boundaries and the class of an entity. Inside, Outside, Beginning (IOB) is one of the most common tagging formats. In IOB, the B- prefix indicates that the token is the beginning of an entity, the I- prefix indicates that the token is inside an entity, and O tag indicates that a token belongs to no entity (Perera et al., 2020). Figure 1 shows an annotated sample for the virus dataset, where UBIAI tool is used for the manual annotation (UBIAI, 2022). We propose that this dataset of 1104 annotated abstracts now offers a GSC for training future NER models in the automated extraction of host-pathogen detection methods from scientific publications.

## 3 Transformer-based Deep Neural Model

This section describes the model architectures, training, and evaluation procedures for the named-entity recognition (NER) task we performed. All our code, written in Python & using Google TensorFlow, will be made freely available online upon publication.

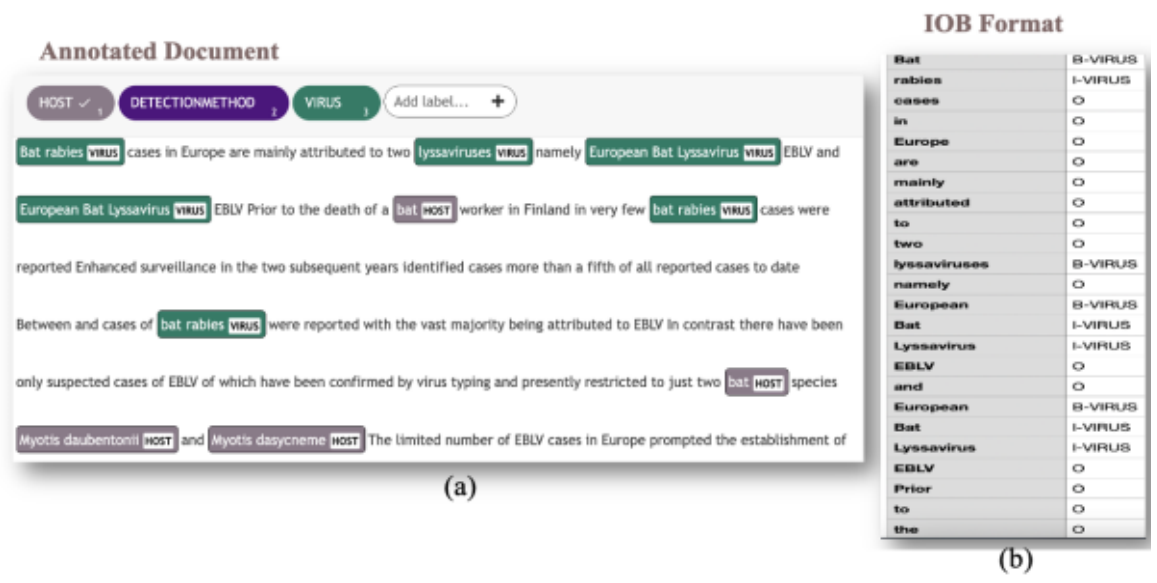

Figure 1: Dataset annotation according to Inside, Outside, Beginning (IOB) cold standard corpus. (a) shows the annotation process that includes highlighting and classifying the entities, (b) shows the output annotated in form of IOB standard.

*Transformer models* were the first deep neural network-based sequence transduction model based entirely on the concept of *attention*. The model architecture is composed of the transformer's *encoder*, based on the original implementation described in (Vaswani et al., 2017), and followed by a classification model. Similar to other sequence processing models, the architecture first uses an embedding layer to convert the input tokens into a feature vector representation and a positional encoding layer to provide information about the order of the sequence. The encoder block consists of self-attention layers, normalization layers, and feed-forward layers (i.e., a multilayer perceptron (MLP)), and outputs a vector for each time step of an input sequence. The classification model uses a feed-forward network to classify these sequences into predefined named entities, therefore performing a *sequence classification task*.

We deployed a *Bidirectional Encoder Representations from Transformers* (BERT) model, a transformer-based model that leverages a fine-tuning-based approach for applying a pretrained language model, i.e., a model trained on a generic task in a semi-supervised manner, and then fine-tuned on a specific task in a supervised manner (Devlin et al., 2018). Leveraging pretrained language models significantly improves performance on many tasks, especially when labeled data is scarce.

Three distinct pretrained BERT models were used, each followed by a classifier model to project the output onto predefined named entities. Since there is no available BERT model that is pretrained on virus and host-related biological literature, available models pretrained on general biological and biomedical literature were used: (1) **BioRedditBERT**, pretrained on large biomedical documents and health-related Reddit posts (Basaldella et al., 2020), (2) **SapBERT**, pretrained on abstracts from PubMed and full-text articles from PubMed Central (Liu et al., 2021), and (3) **Biobert_ncbi_disease_ner**, fine-tuned for NER task on NCBI disease dataset. The NCBI dataset consists of 793 PubMed abstracts and contains 6,892 disease mentions (Doğan et al., 2014). All three models are hosted on the HuggingFace model repository.

The pretrained model may be used as a feature extractor by freezing the model's weights and training only the classification model on the target dataset, or, the weights of some neural layers may be unfrozen and updated on the target task, which is known as *fine-tuning*. Since these models were pretrained on a different corpus, we obtained slightly better results using fine-tuning.

## 4 Results and Evaluation

To evaluate and compare NER models using Gold Standard Corpora, it is required to use standardized evaluation scores. A frequently used error measure is the F-Score, a combination of Recall and Precision. NER models are also evaluated using the accuracy metric. Table 1 shows the evaluation performance of the basic transformer model and performance after fine-tuning of the three pretrained BERT models. Table 2 shows the evaluation performance of the models using the feature extraction learning method described in the previous section. SapBERT obtains the best performance in both fine-tuning and feature extraction learning, probably due to its relatively general nature. Table 3 shows the loss metrics after training the models for 20 epochs.

The visualized annotations in Figure 2 show that the SapBERT model was able to detect and classify almost all the entities of interest: both taxonomic names and detection method names(the latter is a novel result), that appeared in the abstracts.

Table 1: Performance evaluations of the transformer and the three pretrained BERT models finetuned on our novel dataset.

| Model | Accuracy | Precision | Recall | F1-Score |
|---|---|---|---|---|
| Transformer | 0.9826 | 0.9793 | 0.9826 | 0.9795 |
| BioReddit BERT | 0.9814 | 0.9772 | 0.9814 | 0.9770 |
| SapBERT | **0.9857** | **0.9870** | **0.9857** | **0.9853** |
| Biobert_ncbi_disease_ner | 0.9846 | 0.9840 | 0.9846 | 0.9832 |

Table 2: Performance evaluations of the three pretrained BERT models trained on our novel dataset using the feature extraction approach.

| Model | Accuracy | Precision | Recall | F1-Score |
|---|---|---|---|---|
| BioReddit BERT | 0.9655 | 0.9376 | 0.9655 | 0.9508 |
| SapBERT | **0.9849** | **0.9859** | **0.9849** | **0.9844** |
| Biobert_ncbi_disease_ner | 0.9845 | 0.9840 | 0.9846 | 0.9832 |

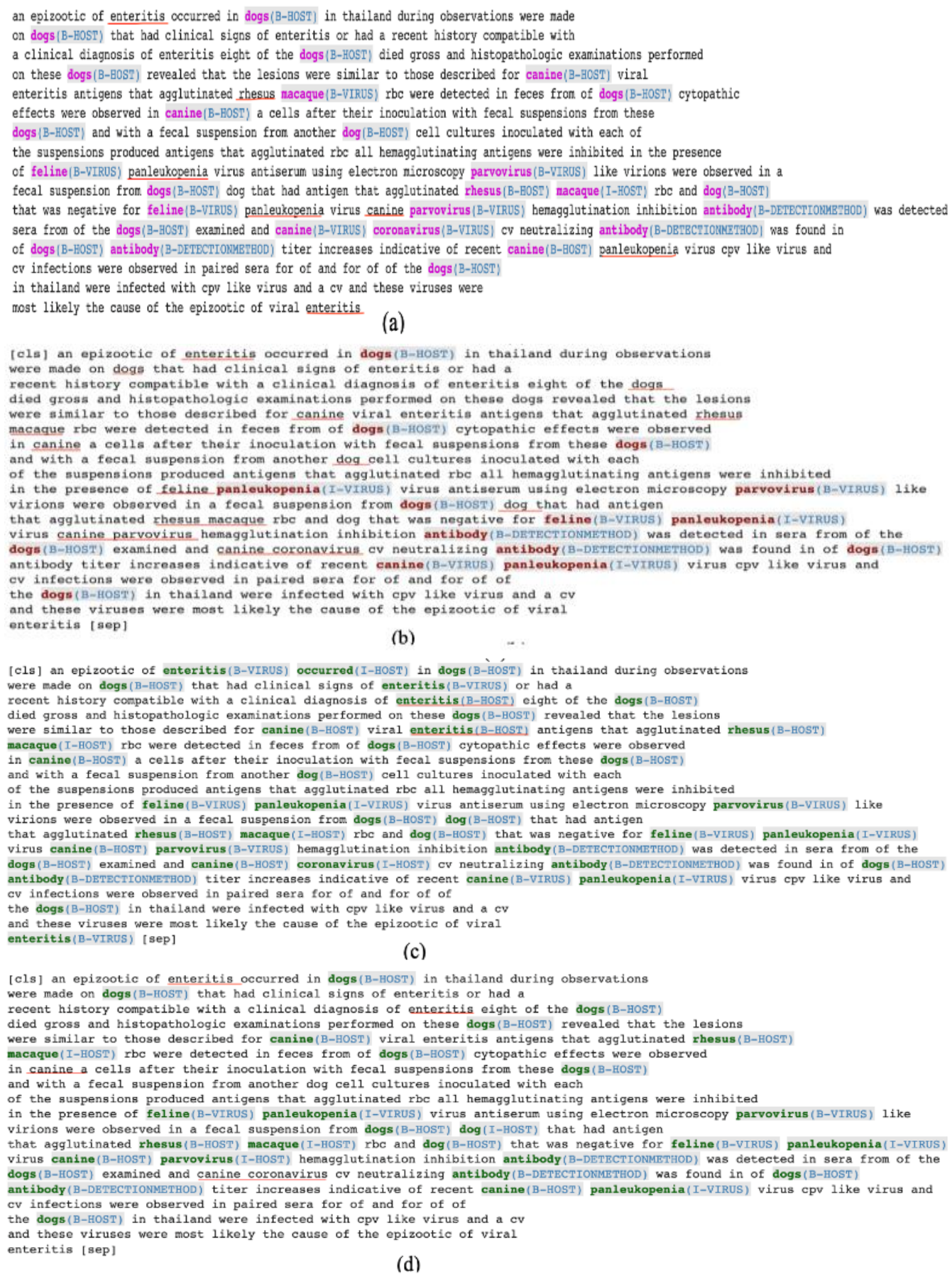

Figure 2: The visual results of the transformer and the three pretrained BERT models: (a) transformer (b) BioRedditBERT (c) SapBERT (d) Biobert_ncbi_disease _ner. All the BERT models were pretrained on biological and health-related literature, and then fine-tuned on our novel dataset. Red lines underscore unrecognized entities in each subfigure.

Table 3: Performance evaluations of the models on the virus dataset

| Model | Loss |
|---|---|
| Transformer | 0.0624 |
| BioRedditBERT | 0.0911 |
| SapBERT | 0.0736 |
| Biobert_ncbi_disease_ner | 0.0904 |

## 5 Conclusions, Impact & Potential

We have presented a novel dataset of significance to the important concept of virus-host association, and therefore to the emergence of pandemics such as the COVID-19 pandemic, and promising initial results on the NER task of identifying both taxonomic names and experimental detection methods. We claim that our dataset of manually annotated abstracts now offers a Gold Standard Corpus for training future NER models in the automated extraction of virus-host and other pathogen detection methods from the biological literature. Several other entities, particularly geographical entities and entities describing species migration, are also relevant to the virus-host association. As a result, immediate next steps will consist of recognizing these entities, and also automatically annotating the full text of the article using semi-supervised methods, in lieu of manually annotating the abstracts.

Recognized taxonomic entities in particular can be linked with knowledge graphs representing taxonomic synonymy as well as more complex taxonomic relationships. These graphs have been used (ATCR, 2022) to reason using automated reasoning and inference techniques such as SMT solving and answer-set programming about relationships expressed in a qualitative spatial logical calculus (such as a form of the region connection calculi), with the goals of resolving taxonomic ambiguity or inferring unspecified relationships. This has been used to align and disambiguate published taxonomies of primates and other species (Franz, N.M. et al., 2016). Further, the approach has the potential to be used in biodiversity conservation applications (Sen, A., Sterner, B., et al., 2021). Such inference may be seen as a generalized form of querying or question-answering over taxonomic graphs, and moreover provides a highly intuitive and visual representation of taxonomic flux over time.

Augmenting these graphs of logical taxonomic relationships with automatically extracted context from the biological literature will have the important benefits of serving to identify novel application domains and providing extra-biological context (e.g., geospatial context) to known & inferred taxonomic relationships.

Further, taxonomic automated reasoning systems have previously been combined (Sen, A., Sterner, B., et al., 2021) with statistical features extracted from biological image repositories (such as citizen-sourced or herbarium-sourced images) to further facilitate the taxonomic relationship discovery task. While we have only considered textual abstracts in our work so far, further useful context may thus be added by augmenting taxonomic knowledge graphs with images or tables extracted from the full text of the publications.

The recognition of a variety of intermediary entities (e.g., locations, methods, migration patterns) is likely to facilitate the discovery of the relevant ecological contexts of the host-virus associations, which, in turn, are subjectively known to be dependent (in some currently undiscovered manner) upon these entities. The extraction of such scientifically informative relationships is a further tangible step ahead.

Finally, these extracted relationships may be considered as background structure for *learning an explainable theory of viral spillover* (from other mammals to humans), when taken together with known examples of such spillover, and known negative examples. Symbolic machine learning techniques such as Inductive Logic Programming (ILP) may be able to exploit such structured data and background knowledge to learn logical relationships that generalize from these data, expressed in a subset of first-order logic and interpretable directly by humans: it is in this sense that we use the term *explainable*.

## Acknowledgment

Research reported in this publication was supported by the National Institute of Allergy and Infectious Diseases of the National Institutes of Health. The award number has been omitted for anonymity.